\documentclass[10pt,twocolumn,letterpaper]{article}

\usepackage[pagenumbers]{cvpr} % To force page numbers, e.g. for an arXiv version

\definecolor{cvprblue}{rgb}{0.21,0.49,0.74}
\usepackage[pagebackref,breaklinks,colorlinks,allcolors=cvprblue]{hyperref}
\usepackage{diagbox}
\usepackage{makecell}
\usepackage{booktabs}
\usepackage{rotating}
\usepackage{multirow}

\title{Unison: A Fully Automatic, Task-Universal, and Low-Cost Framework for Unified Understanding and Generation}

\author{
{Shihao Zhao\thanks{Equal Contribution, $\dagger$Corresponding Author}}$^{*1}$, 
{Yitong Chen}$^{*3}$,
{Zeyinzi Jiang}$^{*2}$,
{Bojia Zi}$^4$, 
{Shaozhe Hao}$^1$, \\ \\
{Yu Liu}$^2$, 
{Chaojie Mao}$^{\dagger2}$, 
{Kwan-Yee K. Wong}$^{\dagger1}$\\ \\
\small $^{1}$The University of Hong Kong \qquad
\small $^{2}$Tongyi Lab, Alibaba Group \qquad
\small $^{3}$Fudan University \qquad
\small $^{4}$The Chinese University of Hong Kong 
}

\begin{document}
\maketitle
\begin{abstract}
Unified understanding and generation is a highly appealing research direction in multimodal learning. There exist two approaches: one trains a transformer via an auto-regressive paradigm, and the other adopts a two-stage scheme connecting pre-trained understanding and generative models for alignment fine-tuning. The former demands massive data and computing resources unaffordable for ordinary researchers. Though the latter requires a lower training cost, existing works often suffer from limited task coverage or poor generation quality. Both approaches lack the ability to parse input meta-information (such as task type, image resolution, video duration, etc.) and require manual parameter configuration that is tedious and non-intelligent.
In this paper, we propose Unison which adopts the two-stage scheme while preserving the capabilities of the pre-trained models well. With an extremely low training cost, we cover a variety of multimodal understanding tasks, including text, image, and video understanding, as well as diverse generation tasks, such as text-to-visual content generation, editing, controllable generation, and IP-based reference generation. We also equip our model with the ability to automatically parse user intentions, determine the target task type, and accurately extract the meta-information required for the corresponding task. This enables full automation of various multimodal tasks without human intervention. 
Experiments demonstrate that, under a low-cost setting of only 500k training samples and 50 GPU hours, our model can accurately and automatically identify tasks and extract relevant parameters, and achieve superior performance across a variety of understanding and generation tasks. Our code is available at \href{https://github.com/ali-vilab/Unison}{https://github.com/ali-vilab/Unison}
\end{abstract}
\section{Introduction}
\label{sec_introduction}

In recent years, significant progress has been made in both the understanding \cite{radford2021learning,li2022blip,liu2023visual,bai2025qwen2,chen2024internvl,dai2023instructblip} and generation \cite{rombach2022high,flux2024,yang2024cogvideox,wan2025,Dalle-3} domains of the multimodal learning field. The understanding task involves models parsing input information and completing tasks like classification, question answering, and summarization, whereas the generation task enables models to proactively create new and meaningful outputs based on given conditions. 
Building a framework capable of unifying both understanding and generation tasks is a highly valuable research topic. Currently, there are two main approaches for this problem: one integrates the two types of tasks into a single transformer model, and the other is a two-stage approach which utilizes pre-trained understanding and generation models and performs fine-tuning for domain alignment.

For the single-model approach \cite{wu2024vila,chen2025janus,team2024chameleon,deng2025emerging}, researchers convert multimodal data into tokens of a unified format via their respective encoders. The model then undergoes training through an auto-regressive paradigm which involves next token prediction. To enhance the output quality of generation tasks, some works also incorporate the diffusion process into this framework \cite{xie2024showo,fan2025unified,deng2025emerging}. 
However, this approach incurs extremely high training costs. It typically requires data on the scale of billions and consumes computing resources equivalent to millions of GPU hours. Even with such substantial investments, current models can often only cover a very limited range of task types. Considering these factors, while it holds great promise in terms of technical prospects, this approach has a high practical threshold and low accessibility for ordinary research teams or individuals with limited resources.

\begin{table*}
  \caption{Comparison of Unison with other methods in terms of framework, training set size, number of task types, and planning capability.}
  \label{table_introduction_comparison}
  \centering
  \setlength{\tabcolsep}{3.5mm}{
  \renewcommand{\arraystretch}{1.2}
  \begin{tabular}{lcccc}
    \toprule
    Method\qquad\qquad & Model Framework & Training Set Size & Number of Task Types & Planning Capability\\
    \midrule
    Show-o \cite{xie2024showo} & 1 Stage & 2B & 6 & No \\
    Janus-Pro \cite{chen2025janus} & 1 Stage & 160M & 3 & No \\
    BAGEL \cite{deng2025emerging} & 1 Stage & 2.6B & 8 & No \\
    Mini-Gemini \cite{li2024mini} & 2 Stage & 2.7M & 4 & No \\
    MetaQueries \cite{pan2025transfer} & 2 Stage & 27M & 6 & No \\
    \midrule
    \textbf{Unison} & 2 Stage & 500K & 12 & Yes\\
    \bottomrule
  \end{tabular}
  }
\end{table*}

The core idea of the two-stage approach \cite{li2024mini,ge2024seed,tong2025metamorph,pan2025transfer} is to adopt pre-trained understanding and generation models, and integrate them into a unified system via fine-tuning for domain alignment. Specifically, understanding tasks directly take the stage-one model’s output as the final result, while for generation, the user’s input is first encoded by the understanding model and then passed as context to the stage-two generation model for final output. 
Although existing two-stage methods have relatively lower training costs, they still have some obvious limitations. First, similar to the single-model solutions, most existing studies cover only a small variety of tasks. Second, the two-stage model is often less competitive to the original dedicated models, and it tends to exhibit unsatisfactory performance on both understanding and generation tasks.

A more critical issue is that existing research has not explored how to understand user intentions. As the variety of tasks increases, with different tasks requiring different hyper-parameters, manually setting task types and their corresponding parameters becomes extremely cumbersome, failing to unlock the model’s full potential. 

In this paper, we propose Unison to address all these challenges. Unison adopts the two-stage approach and it integrates text, image, and video understanding tasks, as well as generation tasks such as text-to-visual content generation, editing \cite{brooks2023instructpix2pix,kawar2023imagic}, controllable generation \cite{zhang2023adding,zhao2024uni}, and IP-based reference generation \cite{ye2023ip,wang2024instantid}, all at an extremely low cost. The entire pipeline is fully automatic. Our model can understand user intentions and extract the required parameters to complete understanding or generation tasks autonomously, eliminating the need for manually setting the modes or parameters. The comparison between Unison and other methods is presented in Table \ref{table_introduction_comparison}.

Specifically, we adopt Qwen2.5-VL \cite{bai2025qwen2} and VACE \cite{vace} as our stage-one (understanding) and stage-two (generation) models respectively. For stage one, we construct a planning dataset and perform LoRA \cite{hu2021lora} fine-tuning, enabling the Vision-Language Model (VLM) to understand user intentions and conduct planning for different tasks and parameters. If the task is identified as a generation task, we project the output of the stage-one VLM into an embedding space and feed it as context into the stage-two generation model. This enhances the generation model’s understanding of user intention, achieving accurate generation results across various input scenarios. The entire training process only involves LoRA fine-tuning of the stage-one understanding model for automated planning and training a projector to align the two models, resulting in an extremely low cost. Meanwhile, this framework largely preserves the inherent capabilities of the understanding and generation models. Our contributions are summarized as follows:
\begin{itemize}
    \item We propose Unison which possesses the ability to understand user intentions. This unified understanding and generation model is capable of handling the entire task flow fully automatically, without requiring users to manually set task types or parameters.
    \item Unison features an extremely low training cost while covering understanding tasks across multiple modalities and generation tasks of various categories, making it highly accessible to general researchers.
    \item Experiments demonstrate that our model achieves comparable performance to other popular unified understanding and generation models with significantly higher costs, supported by extensive benchmark comparisons and visualizations.
\end{itemize}
\section{Related Work}
\label{sec_related_work}

\noindent \textbf{Vision-Language Understanding.}
Recent innovations have greatly advanced cross-modal understanding capabilities. CLIP \cite{radford2021learning} aligns textual and visual information through contrastive learning, achieving excellent zero-shot capability. The BLIP \cite{li2022blip,li2023blip,xue2024xgen} series introduces learnable queries and designs multiple losses for domain alignment while enhancing understanding ability. LLaVA series \cite{liu2023visual,liu2024improved} leverages pre-trained Large Language Models (LLMs) and simplifies the VLM training process while reducing costs via instruction tuning. DeepSeek-VL series \cite{lu2024deepseek,wu2024deepseek} focuses on long-text association and visual reasoning tasks, improving logical coherence in multi-turn conversations. Qwen-VL series \cite{bai2023qwen,wang2024qwen2,bai2025qwen2} realizes bidirectional optimization of understanding and generation through an efficient cross-modal fusion architecture and large-scale data pre-training. In this paper, we adopt Qwen2.5-VL \cite{bai2025qwen2} as the stage-one understanding model.

\begin{table*}
  \caption{Overview of tasks supported by Unison.
The first column indicates the input modality, the first row represents the output modality, and each corresponding cell specifies the tasks that Unison can perform under the given input-output combination.}
  \label{table_task_definition}
  \centering
  \setlength{\tabcolsep}{4mm}{
  \renewcommand{\arraystretch}{1.2}
  \begin{tabular}{cccc}
    \toprule
    Input-Output & Text & Image & Video \\
    \midrule
    Text & Text Understanding & Text-to-Image & Text-to-Video\\
    \midrule
    Image & Image Understanding & \makecell{Image Editing \\ Image Controllable Generation \\ Image Reference Generation} & \makecell{Image-to-Video \\ Video Reference Generation} \\
    \midrule
    Video & Video Understanding & / & \makecell{Video Editing \\ Video Controllable Generation} \\
    \bottomrule
  \end{tabular}
  }
  \vspace{-0.8em}
\end{table*}

\noindent \textbf{Vision Generation.}
The development of diffusion models \cite{ho2020denoising,dhariwal2021diffusion,song2020score} has advanced the quality of visual content generation. For image generation, Stable Diffusion \cite{rombach2022high} adopts the CLIP text encoder \cite{radford2021learning} as its language model and U-Net \cite{ronneberger2015u} as its visual model, while FLUX \cite{flux2024} is based on transformers \cite{vaswani2017attention,peebles2023scalable} and designs the network by combining dual-stream and single-stream architectures. For video generation, AnimateDiff \cite{guo2023animatediff} inserts temporal blocks into a pre-trained text-to-image model to generate videos. CogVideoX \cite{yang2024cogvideox}, HunyuanVideo \cite{kong2024hunyuanvideo}, and Wan \cite{wan2025} are all extensively trained based on the DiT \cite{peebles2023scalable} framework and have been open-sourced. Additionally, VACE \cite{vace} is built on Wan and integrates various tasks by organizing video task inputs—such as editing, reference, and masking—into a unified interface. It also possesses excellent single-frame (i.e., image) generation capability. This paper adopts VACE as the stage-two generation model.

\noindent \textbf{Unifying Understanding and Generation.}
For unifying understanding and generation, one approach adopts a single model and maps data of various modalities to tokens at both input and output layers. It then uses autoregressive methods for next-token prediction. Some works are trained based on pre-trained LLMs \cite{wu2024vila,qu2025tokenflow,chen2025janus}, while others are trained from scratch \cite{team2024chameleon,wang2024emu3}. Additionally, to improve the generation quality of visual content, some studies incorporate the diffusion process into the model \cite{xie2024showo,zhou2024transfusion,fan2025unified,shi2024lmfusion,deng2025emerging}. This approach requires an extremely high training cost. As tasks grow more complex, the training costs increase drastically. How to integrate more modalities and tasks also remains an open question. Another approach adopts a two-stage scheme in which stage one uses a VLM model for understanding tasks and stage two leverages a generative model for generation tasks based on the output of stage one \cite{wu2024next,li2024mini,ge2024seed,tong2025metamorph,zhuang2025vargpt,pan2025transfer}. The two stages undergo fine-tuning for alignment to ensure generation consistency. Such methods have lower costs but still suffer from limitations such as a narrow range of task types and insufficient understanding or generation quality. Our Unison adopts the two-stage approach. It covers diverse tasks at an extremely low cost, and most importantly, can understand user intentions to automate the entire understanding or generation process.
\section{Method}
\label{sec_method}

Unison adopts a two-stage approach. In stage one, a pre-trained VLM is utilized for understanding, which is referred to as the understanding model. In stage two, a pre-trained generative model is employed for content generation, which is referred to as the generation model. 
Specifically, to achieve full automation, we construct planning data to fine-tune the understanding model, enabling it to decouple the task and parameter information from user inputs. If the instruction is classified as an understanding task, the output of the stage-one understanding model is directly used as the final result. If the instruction is classified as a generation task, the decoupled information forms the hyper-parameters for the stage-two generation model to drive the generation process. To enhance cross-stage integration, we further leverage a trainable projector module to bridge the understanding and generation models for better alignment. 
In the following subsections, we will elaborate our data composition, framework structure, and training methodology in detail.

\begin{figure*}[t]
\centering
  \includegraphics[width=1\linewidth]{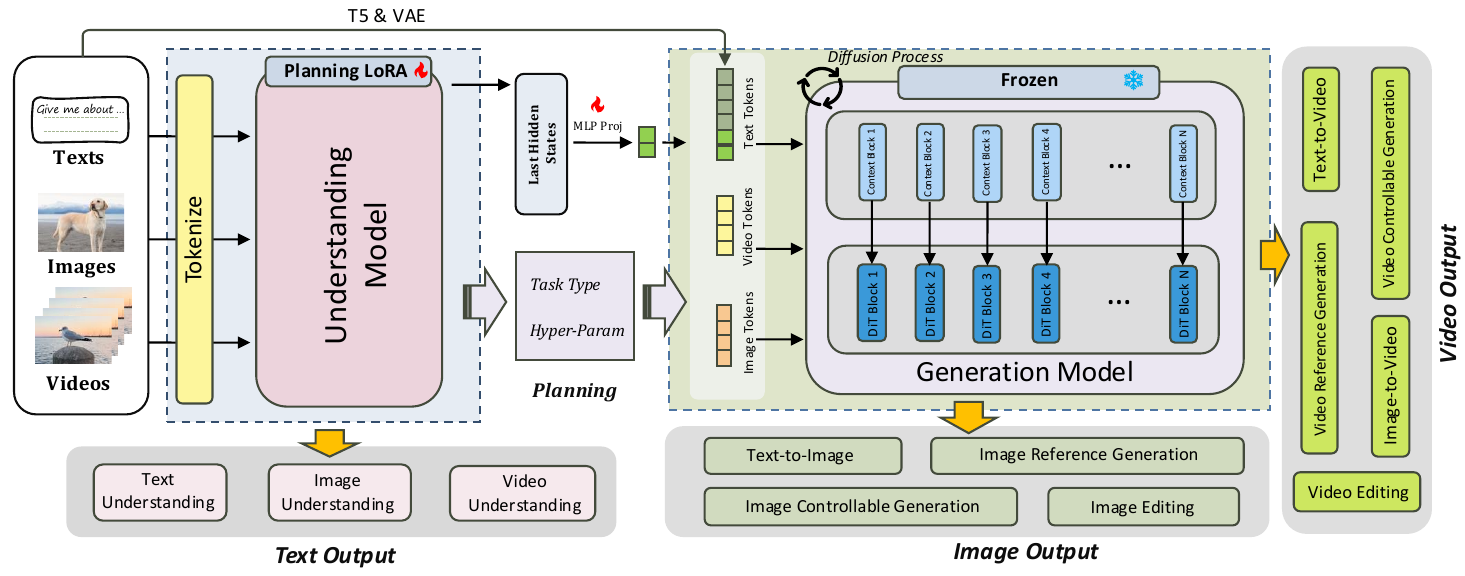}
  \caption{Overview of the Unison. The left shows the stage-one understanding model, the right shows the stage-two generation model. These two components are connected via a projector. For understanding tasks, the stage-one model directly outputs the results. For generation tasks, the stage-one model extracts the task type and the corresponding hyper-parameters from the user's input, then passes this information to the generation model to produce visual content. On one hand, we train the understanding model with LoRA to endow it with the ability to comprehend user intent; on the other hand, we freeze the models in both stages and train the projector for alignment.}
  \label{figure_framework_design_and_training}
\end{figure*}

\subsection{Task Definition}
\label{subsec_task_definition}

We cover a variety of multimodal understanding tasks and multiple generation tasks for visual information, totaling 12 tasks. Specifically, the understanding tasks include text, image, and video comprehension. The generation tasks encompass text-to-visual content generation, editing, controllable generation, and IP-based reference generation, where the visual content covers both images and videos. The detailed task information is listed in Table \ref{table_task_definition}.

\subsection{Data Construction}
\label{subsec_data_construction}

The training data for the stage-one understanding model consist of two parts, namely planning data and ordinary multimodal understanding data. For the second-stage model, we collected diverse generative task data, including images and videos along with their corresponding text, masks, and other relevant information.

\noindent \textbf{Planning Data.}
Planning data are designed to enable the stage-one understanding model to determine whether the user’s intent is to perform a generation task. If so, the model must explicitly indicate the specific type of generation task and the parameters involved in the corresponding task. The format of this data is: input: user instruction; output: $<$Signal Token Start$>$Info$<$Signal Token End$>$. For the generation tasks mentioned in Section \ref{subsec_task_definition}, we have summarized 8 types of signal tokens. Info (if applicable) refers to the parameters associated with the specific task. Below are the specifications of these 8 types of tokens.

\begin{itemize}
    \item \textit{\textless CFI\textgreater: Denotes launching image generation tasks.}
    \item \textit{\textless CFV\textgreater: Denotes launching video generation tasks.}
    \item \textit{\textless BORES\textgreater Width,\ Height\textless EORES\textgreater: Specifies the resolution of the generated image or video.}
    \item \textit{\textless BONF\textgreater Number\ of\ Frames\textless EONF\textgreater: Specifies the total number of frames of the generated video.}
    \item \textit{\textless BOFIDX\textgreater First/Last\ Frame\textless EOFIDX\textgreater: In Image-to-Video task, specifies whether the image uploaded by the user is the first frame or the last frame.}
    \item \textit{\textless BOEDIT\textgreater Mask\ ID,\ Source\ ID\textless EOEDIT\textgreater: In editing tasks, specifies which uploaded image is the mask and which is the image to be edited.}
    \item \textit{\textless CTRL\textgreater: Denotes controllable generation tasks.}
    \item \textit{\textless REF\textgreater: Denotes IP-based reference generation tasks.}
\end{itemize}

For the construction of planning data, we adopt a pipeline of first analyzing categories, then constructing templates, and finally performing random merging to ensure data generalization. Specifically, based on ordinary generative data, we analyze categories for different tasks or parameters. For example, regarding resolution parameters, users may explicitly specify the exact resolution in the input, or only indicate aspect ratio, or provide resolution codes such as 720P and 4K. After listing these scenarios, we use LLMs (Qwen2.5-7B-Instruct \cite{qwen2.5}) to construct diverse templates for each case separately, and finally leverage LLMs to randomly and diversely merge the templates with prompts or instructions from ordinary generative data.

To reduce the training cost, for data where the input contains visual information such as editing or reference generation, the visual part is replaced with \textless Pad Token\textgreater. Therefore, planning data only include textual information rather than visual information. This approach not only effectively preserves the original understanding capability of the stage-one model but also greatly improves training efficiency.

\noindent \textbf{Other Data.}
In addition to planning data, for stage one, we introduce ordinary multimodal understanding data to prevent the understanding model from overfitting to the planning data. This data includes understanding for three modalities: text-text pairs, image-text pairs, and video-text pairs, covering a wide variety of understanding tasks.

For the training data in stage two, we follow the data construction pipeline demonstrated in VACE. Beyond conventional text-to-visual content data, for editing tasks, we use SAM2 \cite{ravi2024sam} to obtain instance-level information and construct data by randomly masking instances. For controllable generation, we extract information such as depth \cite{ranftl2020towards}, scribble \cite{chan2022learning}, grayscale, pose \cite{cao2019openpose}, and optical flow \cite{teed2020raft}. For reference generation, we extract faces or object instances and apply offline or online augmentation operations to create paired data.

\begin{figure*}[h]
\centering
  \includegraphics[width=1\linewidth]{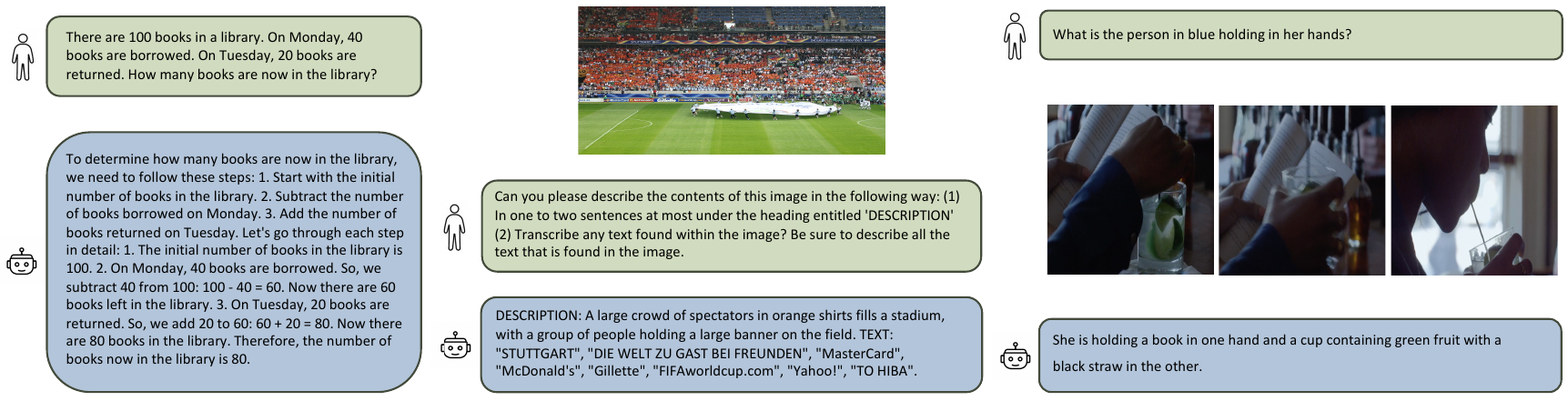}
  \caption{Visualizations of Unison’s multimodal understanding capabilities. From left to right, the results correspond to understanding of text, images, and videos, respectively. In the figures, the green dialogue boxes represent user inputs, and the blue dialogue boxes represent the model’s responses.}
  \label{figure_understanding_qualitative_results}
\end{figure*}

\subsection{Framework Design and Training}
\label{subsec_framework_design_and_training}

We select Qwen2.5-VL \cite{bai2025qwen2}, which possesses diverse multimodal understanding capabilities, as our stage-one understanding model. Meanwhile, we adopt VACE \cite{vace}, a model capable of handling various image and video generation tasks, as our stage-two generation model. Beyond endowing the understanding model with planning capabilities, we only need to align these two models without training additional new capabilities. This fully leverages the rapid advancements in the respective fields of understanding and generation, significantly reducing the training cost of the unified framework. Figure \ref{figure_framework_design_and_training} shows the overall framework of Unison.

\noindent \textbf{Training for Planning.}
Specifically, for Qwen2.5-VL, we fine-tune it using LoRA on the planning data and understanding data detailed in Section \ref{subsec_data_construction}. This not only equips the model with planning capabilities, but also preserves its original understanding capabilities, avoiding overfitting to the planning data. If the stage-one output does not contain signal tokens indicating generation tasks (e.g., \textless CFI\textgreater or \textless CFV\textgreater), it is deemed an understanding task and the output of Qwen2.5-VL is directly used as the result. Otherwise, it is classified as a generation task, and all signal tokens included in Qwen2.5-VL’s output are fed into stage-two as hyper-parameters for VACE to perform generation.

\noindent \textbf{Training for Alignment.}
For the generation tasks in stage two, on one hand, we directly input the user’s initial input as context into the text encoder T5 \cite{raffel2020exploring} of VACE. On the other hand, considering that these user inputs contain a wealth of task-related and parameter information, which creates a certain domain gap with VACE’s text input, we need to align the two. We first project the hidden states from the last layer of stage one via a feedforward layer, and then concatenate the projected states with the T5 embeddings of the user input at the token level. Specifically, the feedforward layer consists of three linear layers, which project the hidden states into a fixed number of tokens, which are then concatenated with the T5 tokens. The concatenated tokens are finally fed into VACE as context. Using the stage-two data described in Section \ref{subsec_data_construction}, we freeze Qwen2.5-VL and VACE and only train the projection layer, thereby reducing the cost while bridging the domain gap. 
\begin{figure*}[t]
\centering
  \includegraphics[width=0.92\linewidth]{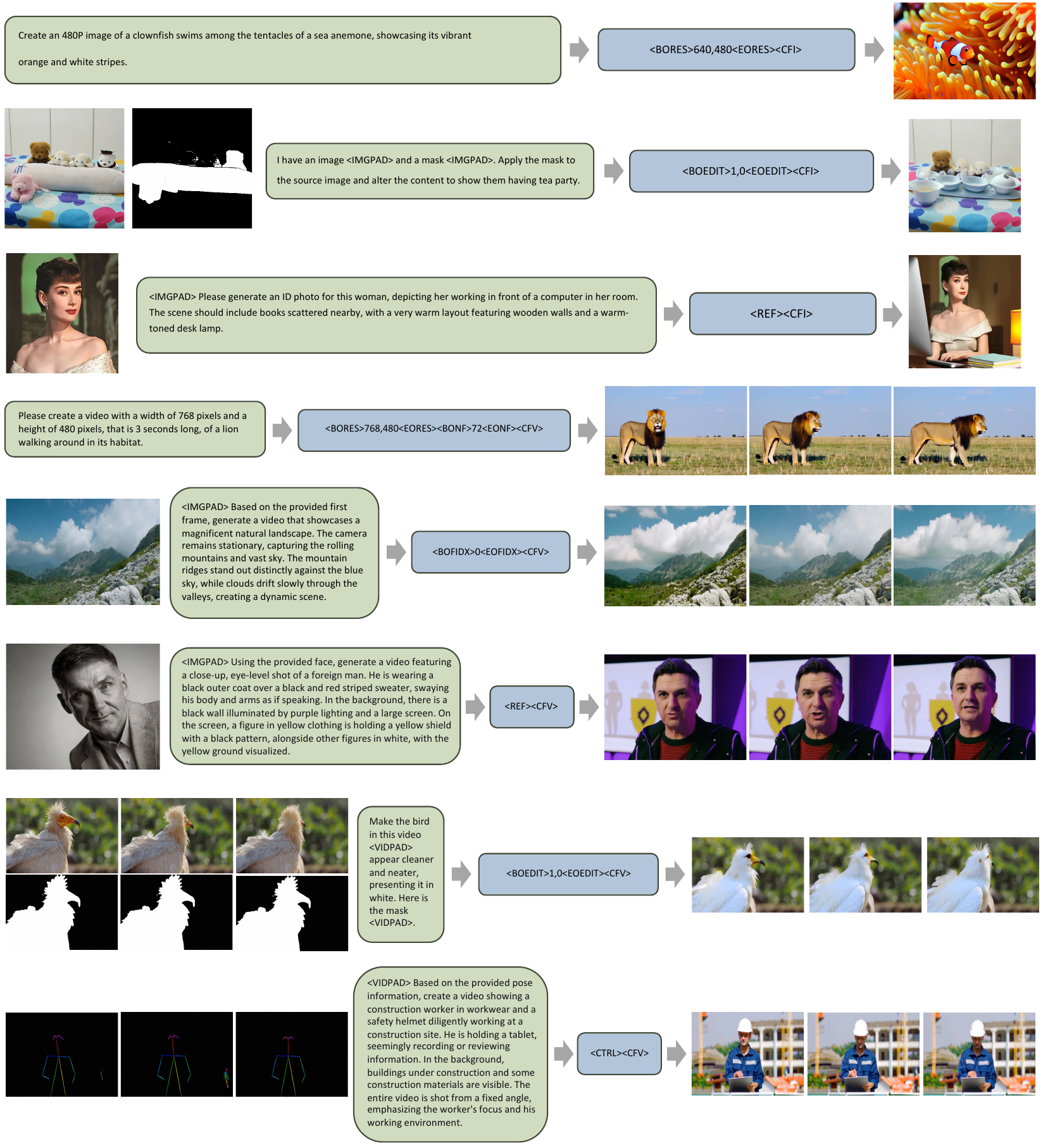}
  \caption{Visualizations of Unison’s multimodal generation capabilities. The leftmost part shows the user input, where the green box contains the input prompt, and the content to the left of the box represents the image, video, or mask condition. The middle blue box shows the output of the stage-one model, which mainly includes signal tokens that guide the generation task of the stage-two model. The right side displays the final generated results. The first three rows correspond to the text-to-image, image editing, and image reference generation tasks, respectively. The following five rows illustrate the text-to-video, image-to-video, video reference generation, video editing, and video controllable generation tasks, respectively. Note that the generated videos are uniformly sampled with three frames for visualization.}
  \label{figure_generation_qualitative_results}
  \vspace{-0.8em}
\end{figure*}

\section{Experiments}
\label{sec_experiments}

In the experimental section, we describe the experimental settings in Section \ref{subsec_experimental_settings}, present the performance of Unison in Section \ref{subsec_main_results}, including visual results across multiple understanding and generation tasks as well as evaluation comparisons on several public benchmarks, and explore the impact of different training strategies on stage one and stage two separately in Section \ref{subsec_ablation_studies}.

\begin{table*}[t]
  \caption{Quantitative evaluation of Unison and other unified understanding and generation methods across different benchmarks. We consider four image understanding benchmarks (POPE, MME-P, MMVP, and MMMU), one video understanding benchmark (MMBench-Video), and two image generation benchmarks (GenEval and DPG-Bench). The best results are highlighted in bold.}
  \label{table_quantitative_evaluation}
  \centering
  \setlength{\tabcolsep}{1.3mm}{
  \renewcommand{\arraystretch}{1.3}
  \begin{tabular}{llccccccc}
    \toprule
    \multicolumn{2}{c}{Method} & POPE $\uparrow$ & MME-P $\uparrow$ & MMVP $\uparrow$ & MMMU $\uparrow$ & MMBench-Video $\uparrow$ & GenEval $\uparrow$ & DPG-Bench $\uparrow$ \\
    \midrule
    \multirow{7}{*}{\rotatebox{90}{Uni. 1 Stage}} 
                      & Show-o \cite{xie2024showo} & 80.0 & 1097.2 & - & 26.7 & - & 0.53 & -\\ 
                      & VILA-U \cite{wu2024vila}  & 85.8 & 1401.8 & - & - & - & - & -\\
                      & TokenFlow-13B \cite{qu2025tokenflow}  & 86.8 & 1545.9 & - & 38.7 & - & 0.55 & 73.38\\
                      & Janus-Pro-1B \cite{chen2025janus}  & 86.2 & 1444.0 & - & 36.3 & - & 0.73 & 82.63\\
                      & Janus-Pro-7B \cite{chen2025janus}  & 87.4 & 1567.1 & - & 41.0 & - & 0.80 & \textbf{84.19}\\
                      & BAGEL-1.5B \cite{deng2025emerging}  & - & 1610.0 & 54.7 & 43.2 & - & - & -\\
                      & BAGEL-7B \cite{deng2025emerging}  & - & \textbf{1687.0} & \textbf{69.3} & \textbf{55.3} & - & \textbf{0.82} & -\\
                      
    \midrule
    \multirow{4}{*}{\rotatebox{90}{Uni. 2 Stage}}
                      & Mini-Gemini-7B \cite{li2024mini} & - & 1523.0 & - & 36.1 & - & - & -\\
                      & VARGPT-9B \cite{zhuang2025vargpt} & 84.4 & 1488.8 & - & 36.4 & - & - & -\\
                      & MetaQuery-L \cite{pan2025transfer} & - & 1574.3 & - & 53.1 & - & 0.78$^\dagger$ & 81.10\\
                      & \textbf{Unison} & \textbf{88.3} & 1543.0 & 67.3 & 35.0 & \textbf{0.94} & 0.70 & 81.01\\
    \bottomrule
  \end{tabular}
  }
\end{table*}

\subsection{Experimental Settings}
\label{subsec_experimental_settings}

\begin{table*}
  \caption{Ablation study on the training method for the stage-one model. The results are presented for five understanding task benchmarks: POPE, MME-P, MMVP, MMMU, and MMBench-Video, along with the GPU hours required for training. The best results are highlighted in bold.}
  \label{table_fine-tuning_for_learning_planning}
  \centering
  \setlength{\tabcolsep}{4mm}{
  \renewcommand{\arraystretch}{1.2}
  \begin{tabular}{lcccccc}
    \toprule
    Method & POPE $\uparrow$ & MME-P $\uparrow$ & MMVP $\uparrow$ & MMMU $\uparrow$ & MMBench-Video $\uparrow$ & GPU Hours $\downarrow$ \\
    \midrule
    Full-Tuning & 87.3 & 1462.4 & \textbf{69.1} & 29.7 & 0.92 & 58 \\
    LoRA & \textbf{88.3} & \textbf{1543.0} & 67.3 & \textbf{35.0} & \textbf{0.94} & \textbf{28} \\
    \bottomrule
  \end{tabular}
  }
  \vspace{-1em}
\end{table*}

\begin{figure*}[h]
\centering
  \includegraphics[width=0.94\linewidth]{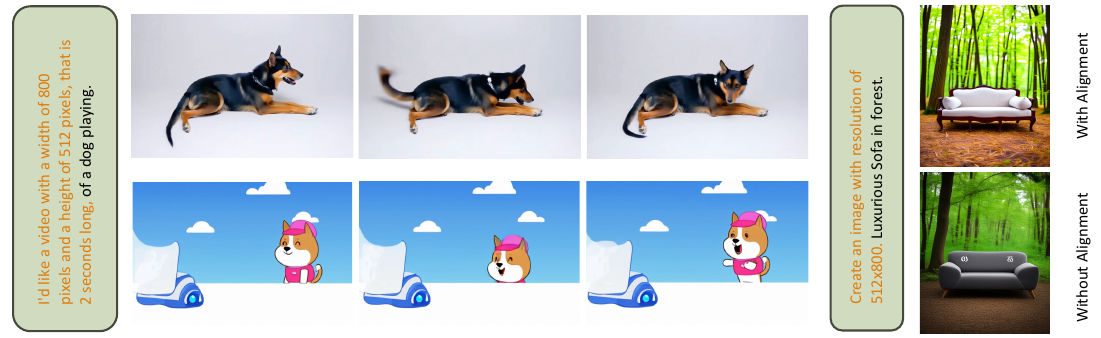}
  \caption{Ablation study on whether alignment is performed between the understanding and generation models in stage two. The left example illustrates a video generation task, and the right example shows an image generation task. In each case, the green box represents the user input, where the orange part indicates the task type and hyper-parameters that are unrelated to the visual content. The first row of results corresponds to using the projector for alignment, and the second row corresponds to no training or alignment being applied.}
  \label{figure_understanding_and_generation_alignment}
  \vspace{-1em}
\end{figure*}

For stage one, we adopted Qwen2.5-VL-3B-Instruct \cite{bai2025qwen2} as the understanding model. The training dataset consists of 200k planning samples, 200k multimodal understanding samples, totaling 400k samples. To endow Qwen2.5-VL with planning capabilities for understanding user intentions while preserving its inherent multimodal understanding abilities, we performed fine-tuning using LoRA. Specifically, we incorporated LoRA into all linear layers, with the LoRA rank set to 32. We use a constant learning rate of $1\times 10^{-5}$ and adopt AdamW optimizer \cite{DBLP:journals/corr/KingmaB14} for optimization. We trained for 1 epoch under this configuration, which took 3.5 hours on 8 A800 GPUs.

For stage two, we employed Wan2.1-VACE-1.3B \cite{vace} as the generation model. The training data covers various types of generative task data mentioned in Table \ref{table_task_definition}, totaling 100k samples. During this stage, we froze Qwen2.5-VL and VACE, and only trained the projection layer which we mentioned in Section \ref{subsec_framework_design_and_training}. We set the learning rate to $1\times 10^{-3}$ and use AdamW optimizer \cite{DBLP:journals/corr/KingmaB14}. We trained the projector for 1 training epoch, which took 2.5 hours on 8 A800 GPUs.

\subsection{Main Results}
\label{subsec_main_results}

In this section, we evaluate the performance of our Unison. On the one hand, we present the visualization results for the aforementioned tasks. On the other hand, we also conduct quantitative comparisons between Unison and other unified understanding and generation methods on various understanding and generation benchmarks.

\noindent \textbf{Qualitative Results.}
We visualize the results of Unison across different understanding tasks in Figure \ref{figure_understanding_qualitative_results} and generation tasks in Figure \ref{figure_generation_qualitative_results}. For the understanding tasks, the stage-one model demonstrates strong abilities in reasoning, recognition, and summarization, etc, accurately answering questions and comprehending content whether the input is text, images, or videos. For the generation tasks, it can be observed that the stage-one model is capable of accurately understanding the user’s intent, correctly identifying the type of task the user intends to perform based on the input, and precisely outputting the corresponding hyper-parameters for that task; for example, in Figure \ref{figure_generation_qualitative_results}, the fourth row (text-to-video task) shows the stage-one model accurately classifying the task as text-to-video and correctly outputting the user-specified resolution and video duration. Moreover, the generation model (stage-two) effectively handles a variety of generation tasks, delivering high-quality visual outputs regardless of whether the input includes only text as a condition (e.g., the first-row text-to-image task), reference conditions (e.g., the third-row and sixth-row results), mask conditions (e.g., the second-row and seventh-row editing results), or layout conditions (e.g., the last-row result), with the stage-two model consistently exhibiting excellent visual quality and accurately following the user’s instructions.

\noindent \textbf{Quantitative Evaluation.}
Table \ref{table_quantitative_evaluation} shows the performance of Unison on public benchmarks. We selected four image understanding benchmarks (POPE \cite{li2023evaluating}, MME \cite{mme2023}, MMVP \cite{tong2024eyes} and MMMU \cite{yue2024mmmu}), one video understanding benchmark (MMBench-Video \cite{fang2024mmbench}), two image generation benchmarks (GenEval \cite{ghosh2023geneval} and DPG-Bench \cite{hu2024ella}) to evaluate its multimodal understanding and generation capabilities. As shown in the table, among these benchmarks, we achieve the best performance on the POPE. Additionally, we also consider quantitative evaluation for video understanding. For the other benchmarks, although Unison does not achieve the absolute best results, it still delivers comparable performance when compared to other state-of-the-art (SOTA) methods. This is particularly noteworthy given that our chosen understanding model has only a 3B parameter size, and the visual generation module of our selected generation model is even smaller at 1.3B. More importantly, Unison is trained under extremely low computational and data costs, yet it can effectively handle a diverse and broad set of tasks spanning both understanding and generation across multiple modalities.

\subsection{Ablation Studies}
\label{subsec_ablation_studies}

In this section, we explore the impact of different training strategies on Unison. On the one hand, we compare the differences in performance and training cost between full fine-tuning and LoRA fine-tuning in the stage-one model. On the other hand, we investigate the effect on generation quality when skipping the alignment between the two models in the stage-two training process.

\noindent \textbf{Fine-tuning for Learning Planning.}
To endow Unison with planning capabilities, we adopt LoRA to fine-tune the stage-one understanding model. Here, we compare the performance of full fine-tuning and LoRA-based training on a mixed dataset containing both the same planning data and normal multimodal understanding data, which is mentioned in Section \ref{subsec_data_construction}. The results are shown in Table \ref{table_fine-tuning_for_learning_planning}. It can be observed that, in terms of performance, LoRA training achieves better results on four out of five benchmarks, indicating that LoRA is less prone to overfitting on planning data compared to full fine-tuning. Additionally, it can be observed that the training time required for LoRA fine-tuning is less than half of that needed for full fine-tuning, making it significantly more efficient.

\noindent \textbf{Understanding and Generation Alignment.}
Since the user input in Unison contains task-related information, such as the task type and corresponding hyper-parameters, there exists a certain gap between this type of input and typical user queries in conventional generative models. To bridge this gap, we introduce a projector to connect the understanding and generation models. Here, we compare the approach of using the projector for alignment with the baseline where the user input is directly fed as context into VACE in the second stage, that is, without any explicit alignment training. The results are shown in Figure \ref{figure_understanding_and_generation_alignment}. We find that, in the video generation example, the task-related description occupies the majority of the prompt, while the actual description of the video content consists of only a few words. This greatly affects the judgment of the generation model. In contrast, after performing explicit domain alignment, both the matching degree and quality of the generated video content are significantly better than those in the non-aligned case. In the image generation case, in the unaligned result, strange symbols appear on the sofa, which is also likely caused by the input domain gap.
\section{Conclusion}
\label{sec_conclusion}

In this paper, we introduce Unison, a novel multimodal framework with planning capability that effectively unifies understanding and generation tasks with minimal training cost and maximal task coverage. Unison adopts the two-stage approach and fully leverages pre-trained models in both the understanding and generation domains. At an extremely low cost, it supports a wide range of tasks, including text, image, and video understanding, as well as various generation tasks such as text-to-visual content generation, editing, controllable generation, and IP-based reference generation, a total of 12 tasks.
More importantly, we constructed a planning dataset, enabling Unison to understand the user’s intent, determine the appropriate task type, and extract the necessary parameters, thereby allowing the entire understanding and generation pipeline to be completed automatically. Experiments demonstrate that with just 500k training samples and 50 GPU hours, Unison achieves superior performance across a variety of understanding and generation tasks, supported by extensive benchmark evaluations and qualitative visualizations. 
These results highlight Unison’s potential as a practical, accessible, and powerful solution for unified multimodal AI systems, paving the way for more intelligent and user-friendly multimodal interactions. And we also believe the concept of planning can be adapted and applied to other methods of unifying generation and understanding tasks.
{
    \small
    \bibliographystyle{ieeenat_fullname}
    \bibliography{main}
}

\clearpage

\twocolumn[
  \begin{center}
    \vspace{-0.5cm}
    {\Large \textbf{Appendix}}
    \vspace{1cm}
  \end{center}
]

\section{Construction of Planning Data}
\label{sec_construction_of_planning_data}

Here we use resolution information as an example to elaborate on the construction of the planning data. We first categorize user inputs regarding resolution into the following four types based on their level of detail:

\begin{itemize}
  \item Width and height specifications, e.g., ``the width is 480 pixels and the height is 640 pixels''
  \item Standard resolution terminology, e.g., ``720P''
  \item Aspect ratio specifications, e.g., ``aspect ratio is 16: 9''
  \item General orientation descriptions, e.g., ``horizontal''
\end{itemize}

We then leveraged Qwen2.5 to generate 100 templates for each category. For the first category, templates like ``This image’s height is $<$Placeholder 1$>$, with the width being $<$Placeholder 2$>$'' were created, with the placeholders randomly assigned height and width values. For the latter three categories, we construct templates such as ``It features a display detail level of $<$Placeholder$>$'', ``The aspect ratio (width to height) of the image is $<$Placeholder$>$'', and ``landscape-wide view'', etc. For these three categories, we pre-defined both the placeholder content and their corresponding concrete resolutions. For example: ``1080P'' corresponds to 1920 $\times$ 1080 pixels,``16: 9'' aspect ratio corresponds to 1280 $\times$ 720 pixels, Horizontal and vertical orientations also have pre-defined pixel dimensions. These concrete resolution values serve as the model outputs, i.e., the groundtruth, for training. For other meta-information, such as the number of frames in a video, we also considered several scenarios: when users specify the video duration, or when they indicate the total number of frames. We then constructed templates following the above pipeline. 

\begin{figure}[h]
\centering
  \includegraphics[width=1\linewidth]{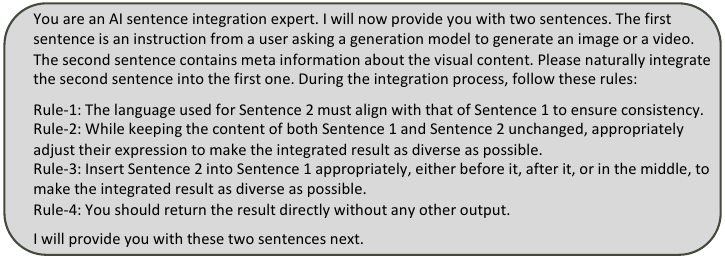}
  \caption{The system prompt used for combining templates and raw user instructions.}
  \label{figure_system_prompt}
\end{figure}

Subsequently, we used Qwen2.5 to randomly combine user inputs that do not contain meta-information, i.e., conventional textual instructions, with these constructed templates to produce the final planning data. Figure \ref{figure_system_prompt} shows the system prompts we employed.

\section{More Visualization Results}
\label{sec_more_visualization_results}

\begin{figure*}[t]
\centering
  \includegraphics[width=1\linewidth]{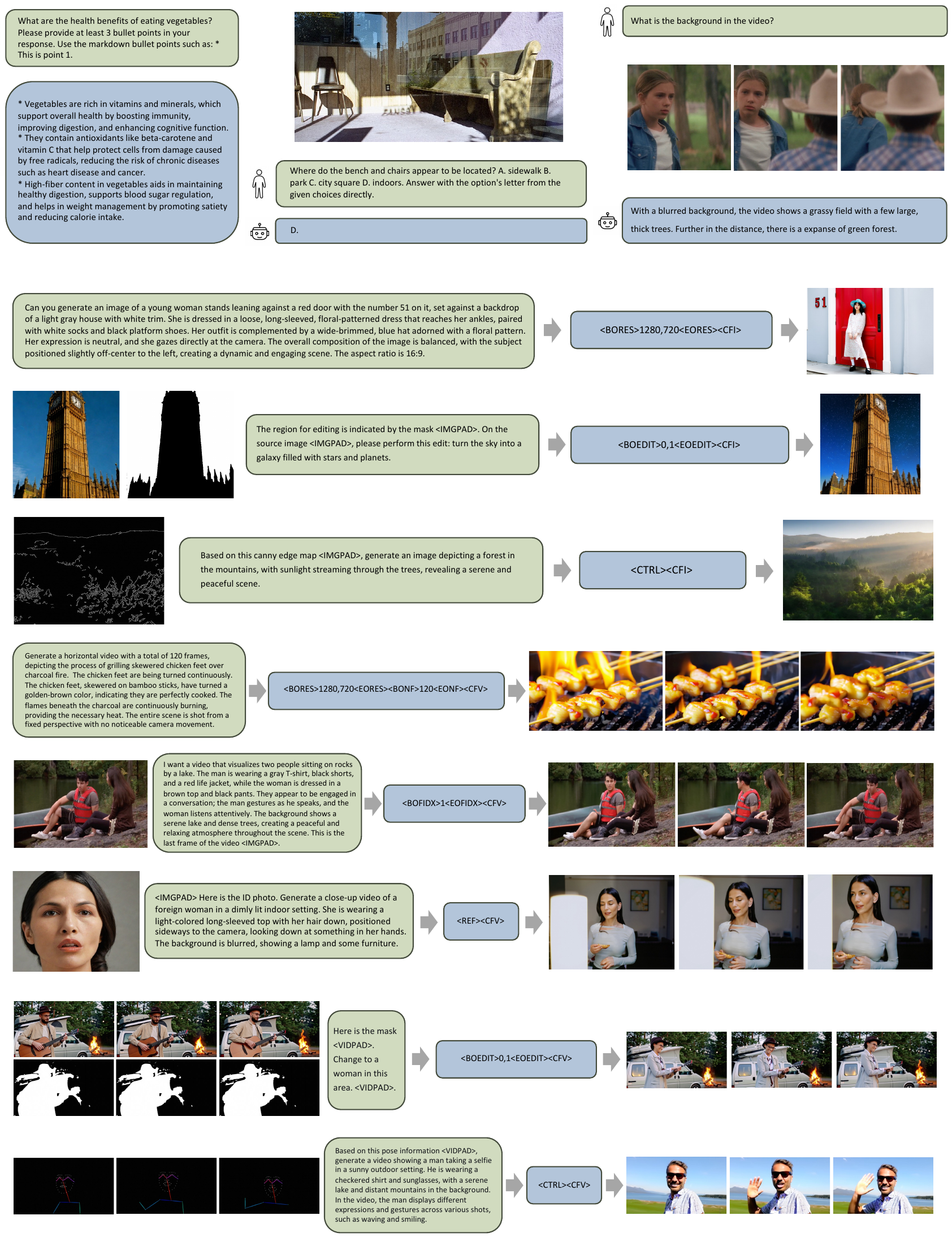}
  \caption{More visualization results on the understanding and generation tasks.}
  \label{figure_more_visualization_results_1}
\end{figure*}

\begin{figure*}[t]
\centering
  \includegraphics[width=1\linewidth]{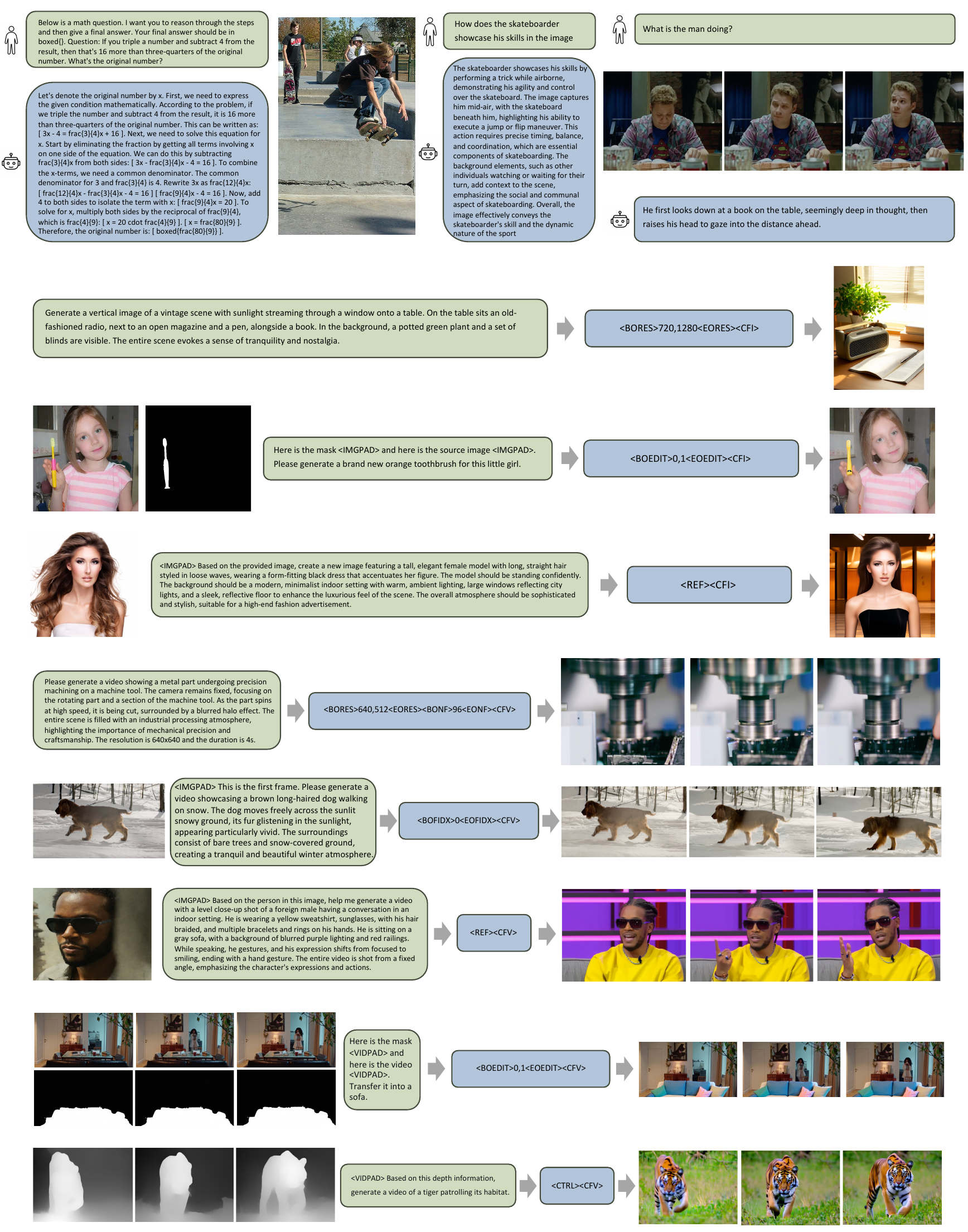}
  \caption{More visualization results on the understanding and generation tasks.}
  \label{figure_more_visualization_results_2}
\end{figure*}

In this section, we present additional visualization results in Figure \ref{figure_more_visualization_results_1} and Figure \ref{figure_more_visualization_results_2}. In each figure, the top three cases demonstrate the model's understanding of text, image, and video inputs. Here, the green boxes contain user inputs, while the blue boxes display the corresponding outputs. The lower cases illustrate tasks related to image and video generation. The leftmost section shows user inputs, where green boxes contain textual prompts, and content to the left of these boxes represents image, video, or mask conditions. The middle blue boxes present outputs from the stage-one model, which primarily consist of signal tokens guiding the generation tasks of the stage-two model. The right side displays the final generated results.

\section{Discussion}
\label{sec_discussion}

Given Unison's low training cost and broad task coverage, we hope it can be conveniently utilized to enhance creative expression and accessibility in positive explorations, while avoiding risks related to misinformation and ethical use. Additionally, we have some technical reflections regarding Unison.

During the training of the understanding model in stage one, we noted that to reduce training cost, the planning data only includes textual information, with all visual content replaced by padding tokens. As a result, the stage-one model can only rely on the user's text input to generate signal tokens and corresponding information. An alternative approach is to train the planning capability directly using raw visual content instead of padding tokens. This would offer certain advantages. For instance, when padding tokens are used, the user must explicitly specify conditions, such as: ``This is the mask $<$Pad Token$>$ and this is the source image $<$Pad Token$>$. Editing instruction''. In contrast, with raw visual content, the user could simply input ``$<$IMG Token$>$, $<$IMG Token$>$. Editing instruction'', and the model could intelligently determine which token corresponds to the mask and which to the source image based on its understanding of the two input images. While this would enhance model intelligence, it would also introduce more complex image or video information compared to text, thereby significantly increasing training cost. Furthermore, although this paper primarily explores a two-stage framework, we believe planning capability could also be effectively applied in one-stage models. We will continue to investigate these questions in future work. 

\end{document}